# A Systematic Analysis for State-of-the-Art 3D Lung Nodule Proposals Generation

Hui Wu, Matrix Yao, Albert Hu, Gaofeng Sun, Xiaokun Yu, Jian Tang

hui.h.wu@intel.com, matrix.yao@intel.com, albert.hu@intel.com,
gaofengx.sun@intel.com, xiaokun.yu@intel.com, jian.j.tang@intel.com

## ABSTRACT

Lung nodule proposals generation is the primary step of lung nodule detection and has received much attention in recent years[1] In this paper, we first construct a model of 3-dimension Convolutional Neural Network (3D CNN) to generate lung nodule proposals, which can achieve the state-of-the-art performance. Then, we analyze a series of key problems concerning the training performance and efficiency. Firstly, we train the 3D CNN model with data in different resolutions and find out that models trained by high resolution input data achieve better lung nodule proposals generation performances especially for nodules in too small sizes, while consumes much more memory at the same time. Then, we analyze the memory consumptions on different platforms and the experimental results indicate that CPU architecture can provide us with larger memory and enables us to explore more possibilities of 3D applications. We implement the 3D CNN model on CPU platform and propose an Intel Extended-Caffe framework which supports many highly-efficient 3D computations, which is opened source at https://github.com/extendedcaffe/extended-caffe.

## KEYWORDS

Lung nodule proposals generation, 3-dimension Convolutional Neural Network (3D CNN), memory consumptions, CPU architecture, Intel Extended-Caffe

## 1  INTRODUCTION

Detecting some malignant nodules from lung cavities by computer-aided diagnosis (CAD) techniques is very important, which can assist in preventing lung cancer deteriorating and improving the cure rate [1-7]. Lung nodule detection can be further divided into two sub-tasks, among which lung nodule proposals generation is the primary step of nodules detection and cancer prediction. The missing identification of true nodules will decrease the probability of preventing lung cancer, while the generation of too many false positives will increase the successive efforts of non-nodules filtering.

To generate lung nodule proposals from CT Scans, some 3-dimension Convolutional Neural Networks (3D CNNs) based methods have been proposed, which apply models of 3D CNN to capture spatial features of lung nodules and achieve high performances. Rotem Golan et al. [8] applied deep CNNs to capture nodule features of some 3D sub-volumes cropped from the input CT image. Huang X et al. [9] leveraged both a priori knowledge and the deep features learned by 3D CNN model for lung nodule detection. Even though 3D CNN based methods have achieved high lung nodule proposals generation performances, there still exists some key problems to be solved.

Firstly, data resolution should be considered in 3D CNN. Nodules in lung cavities are in nearly round shapes with different diameters, which are too small to be detected [10]. Taking data in Luna'16 dataset for example, some nodules (e.g. diameters ranging from 3mm to 30mm) take one millionth of the whole CT scans (e.g. sizes equal to 512mm×512mm×512mm). Capturing effective features of these too small nodules from the much larger sizes of CT scans is so difficult that data resolution should be considered to make the smaller nodules being able to be recognized by the CNN models.

Then, hardware memory consumptions should be considered in 3D CNN. The CNN model has been converted from 2D to 3D, and all the input and output data are in 3D forms, which consume much more memory. Besides, as mentioned above, to explore the impact of data resolution on the performance of 3D CNN model, hardware platforms with large memory which is able to receive high resolution data are in requirement.

Moreover, model training efficiency should be considered. As we know, CNN training is a time-consuming work, and the 3D CNN model consumes much more time to be trained due to its larger amount of parameters and the higher dimension of input data. Thus, training efficiency of 3D CNN model should be improved, making large amount of optimization engineering work being in necessary requirement.

In this paper, by referring to the common philosophy of prior networks [11-13], we construct a 3D lung nodule proposals generation network and implement it on CPU platform using our proposed Extended-Caffe framework, which has achieved state-of-the-art lung nodule proposal generation performance on Luna'16 competition [14]. Based on this 3D CNN model, we analyze the aforementioned key problems in detail, including:

(1) To improve lung nodule proposals generation performance and make the too small nodules be conspicuous, we try to reduce sampling distances on the CT raw data to generate high resolution data and we find out that models trained by higher resolution input data achieve better lung nodule proposals generation performances.
(2) The input data is in 3D form with high resolution, which consumes large hardware memory. We conduct some experiments to analyze memory consumption situations of the 3D CNN model on CPU and GPU, respectively, the results indicating that CPU architecture can provide larger usage memory for 3D models.
(3) For the requirement of large memory, we implement the 3D CNN model on the CPU platform and use the proposed Intel Extended-Caffe framework which has been optimized on Intel MKL-DNN and supports many highly-efficient 3D computations.

## 2 3D CNN based Nodule Proposals Generation

In this section, we systematically analyze lung nodule proposals generation based on 3D CNN, and elaborate some related key problems including data preprocessing, network design, network training, as well as network implementation and framework optimization.

### 2.1 Data Preprocessing

Raw CT scans contains a sequence of 2D images, and the interval between these 2D images is called Z-interval, which are measured on millimetre basis. Usually, the intervals of raw CT scans obtained from different CT instruments are different. Taking CT scans in Luna'16 dataset [14] for example, Z-interval of the CT scans in this dataset range from 0.625mm to 2.5mm, which are the same for the X- and Y- intervals. In order to make a deep learning model work on a unified dataset, we have to interpolate the pixels of the original raw CT in X, Y, and Z directions by taking a fixed sampling distance, so that a raw CT is converted to a new 3D image where the pixel-to-pixel intervals in three directions equal to the sampling distance. Different sampling frequencies result in different sizes of CT data and different diameters of nodules. We respectively set sampling distance as 1mm, 1.33mm and 2mm, and show the sampling results in Table 1, from which we can see that for the same CT scan, small sampling distance will generate higher resolution data and larger nodule sizes. Thus, in this paper, we set the sampling distance as 1mm to generate relatively larger sizes of nodules. Some experiments are conducted in Section 3 to quantitatively analyze the impact of data resolution on lung nodule proposal generation performance.

**Table 1: Example of sampling results with different sampling distances.**

| Sampling distance (mm) | Data size (pixel) | Nodule size (pixel) | Crop size (pixel) |
|---|---|---|---|
| 1.00 | 249×256×302 | 3.66 | 128×128×128 |
| 1.33 | 188×196×231 | 2.74 | 96×96×96 |
| 2.00 | 127×136×159 | 1.83 | 64×64×64 |

Note, unlike the ordinary object detections, lung nodule detections suffer a unique problem that a nodule takes only one millionth of the whole CT as shown in Table 1. Therefore, in order for a model to effectively extract the features of nodules, we crop small sub-regions (i.e. crops) from the CT data and then feed these crops into the model for training. Again, smaller sampling distances lead to bigger crops.

### 2.2 Network Design

*2.2.1 Network Structure.* By referring to the common philosophy of prior networks [11-13], we construct a 3D CNN model for lung nodule proposals generation and the main structure is shown in. Fig. 1, which is separated into down-sampling and up-sampling parts. The down sampling part contains five groups of 3D residual blocks interleaved with four pooling layers. Each residual block consist of convolution, batch-norm, ReLU and some other operations together with a residual structure (C1 and C2 in Fig. 1). On the other hand, the up-sampling is done through two de-convolutions (Deconv in Fig. 1), and we combine the output of each deconvolution with the output of the corresponding down-sampling layer, so as to generate the feature maps that contains both local and global information of the original input data.

*2.2.2 Loss.* The network has two sibling output layers. The first outputs a two class labels, which indicates the corresponding proposal being a nodule or not. The second sibling layer outputs 3D bounding-box regression offsets, indicating by the center ($x$, $y$, $z$) and the length $d$. For each labeled nodule, a multi-task loss $L$ is used to jointly train for classification and bounding-box regression:

$$L(p,g,v,t) = L_{cls}(p,g) + L_{loc}(t,v) \quad (1)$$

where $L_{cls}(p,g)$ is the classification loss computing by binary cross entropy between the predicted label $p$ and the ground truth label $g$. $L_{loc}(t,v)$ is the regression loss computing by smooth $L1$ loss between the predicted location $t=(t_x,t_y,t_z,t_d)$ and the ground truth location $v=(v_x,v_y,v_z,v_d)$, respectively.

$$L_{loc}(t,v) = \sum_{i \in \{x,y,z,d\}} \text{smooth}_{L1}(t_i - v_i) \quad (2)$$

$$\text{smooth}_{L1}(x) = \begin{cases} 0.5x^2 & \text{if } |x|<1 \\ |x|-0.5 & \text{otherwise,} \end{cases} \quad (3)$$

Three anchors of different scales (10mm, 30mm and 60mm with 1mm resolution) are set at each point on the output feature map. For each crop ($m \times m \times m$), the network generates output vector with dimension of $\frac{m}{4} \times \frac{m}{4} \times \frac{m}{4} \times 3 \times 5$, where $m$ is the size length of the input crops.

### 2.3 Network Training

We implement the designed network on CPU platform using proposed Extended-Caffe framework which will be elaborated in Section 2.4. Stochastic gradient descent (SGD) and step-wise learning rate strategies are adopted to train the network. All network parameters are initialized randomly, and the initial learning rate is set as 0.01. We train the network for 100 epochs, when the epoch value reaches 50 and 80, the learning rate is reduced by 0.1, respectively.

Taking the training process of subset 0 in Luna'16 dataset for example, loss curves over the training epochs are shown in Fig. 2, from which we can see that the network training losses are reducing over the epochs and remain stable after 60 epochs. The final total loss of the network is close to 0.1.

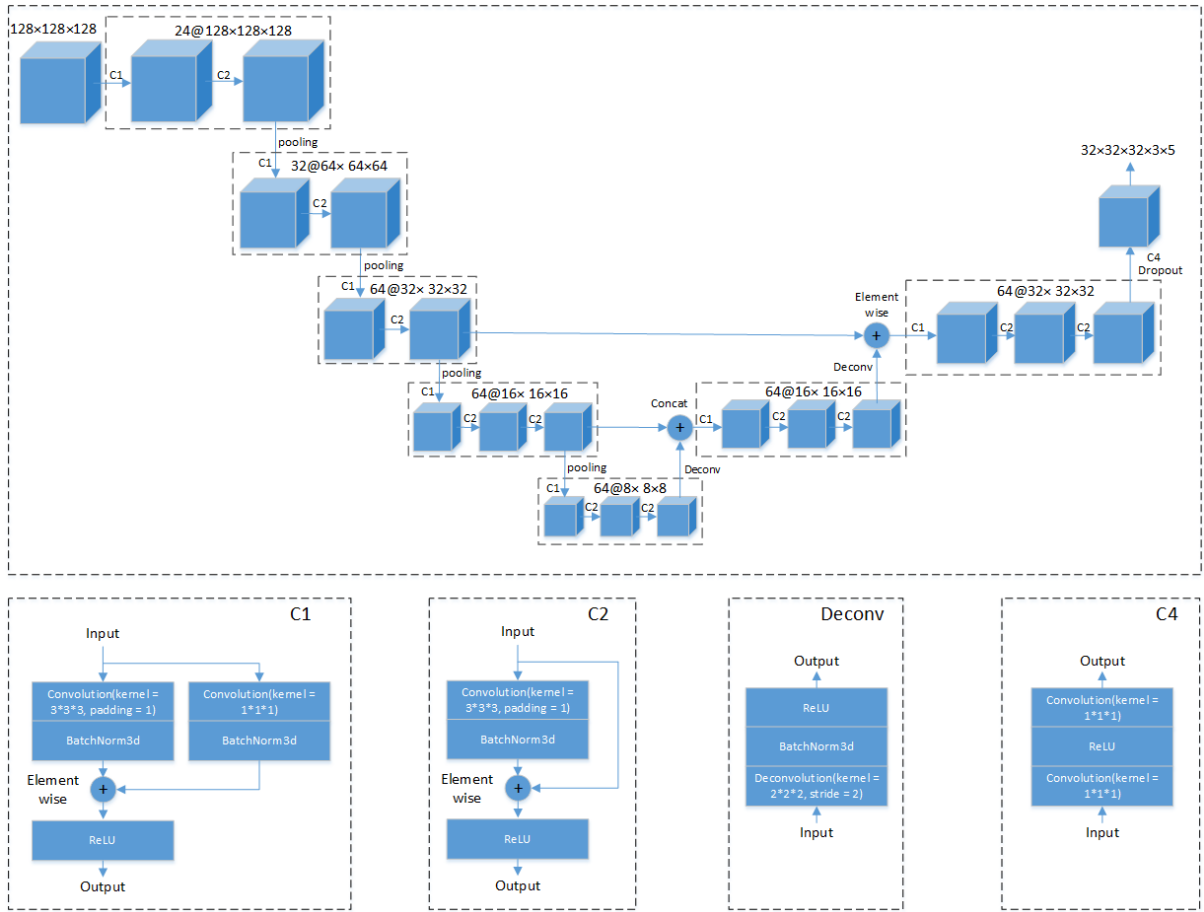

**Figure 1: Nodule proposal generation network architecture.**

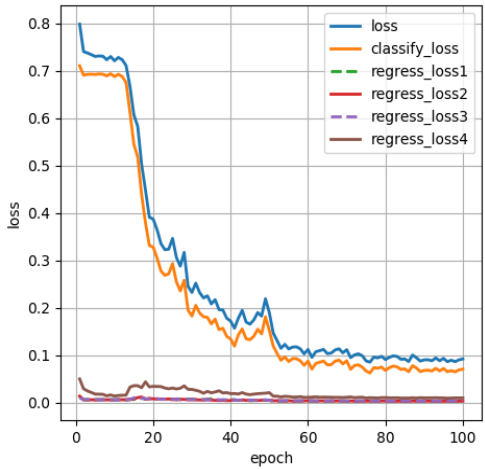

**Figure 2: Example of training loss curves.**

## 2.4 Network Implementation and Optimization

To meet the requirement of large memory, we implement the 3D CNN model on CPU platform and use Caffe framework [24]. Based on the official Caffe, we implement some frequently used 3D computations and further optimize the 3D convolution and 3D batch norm units.

*2.4.1 3D Convolution Optimization.* Convolution is the most commonly used operation in CNN. In this paper, we propose two strategies of implementing 3D convolutional functionalities and compare their performances in Section 3.

The official Caffe implements 2D convolution based on IM2COL and SGEMM strategies, the principle of which is illustrated in Fig. 3(a). Firstly, both the data and kernel are rearranged to columns, and then SGEMM process is performed between the rearranged kernel and data. Combining IM2COL and SGEMM strategies, convolutional operation is converted into multiplication between matrixes, which can accelerate convolutional computation. Based on this principle, we implement the 3D functionality in the same way.

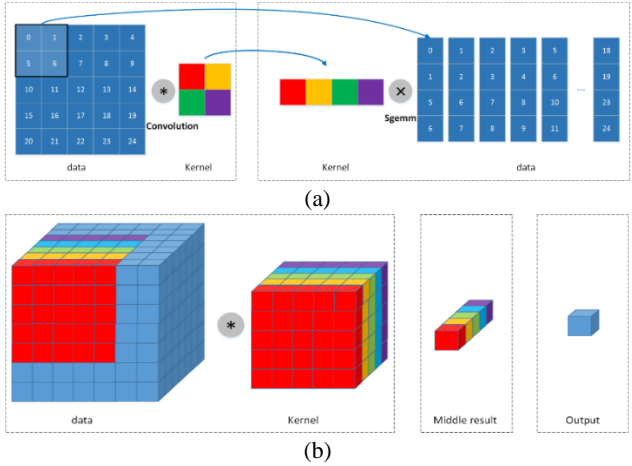

Figure 3: The principles of 3D convolution based on different strategies. (a) SGEMM based method, (b) MKL-DNN based method.

Moreover, we also implement 3D convolution based on Intel Math Kernel Library for Deep Neural Networks (MKL-DNN) [15], which is an open source performance library for Deep Learning (DL) applications intending to accelerate DL frameworks on Intel architecture. 2D convolution in MKL-DNN on CPU platform is highly accelerated based on Intel AVX512 architecture instruction set which represents a significant leap to 512-bit Single Instruction Multiple Data (SIMD) support, and can offer higher performance for the most demanding computational tasks. We take this advantage of Intel CPU platform and implement the 3D convolution based on the 2D version in MKL-DNN, as illustrated in Fig. 3(b), which can be treated as a series of element wise multiplications followed by summations. The input of 3D data can be treated as a group of 2D slices. Each slice will be convolved by a 2D slice of the 3D kernel to get series of middle slices which corresponds to a specific output slice and will be summed to get the final output feature map.

*2.4.2 3D Batch Norm Optimization.* As recommended in the official Caffe, two layers of scale and batch norm are always used together during the network construction, which are implemented separately. In Intel Math Kernel Library (MKL) 2017 [16], a batch norm layer will be merged with its successive scale layer, which can significantly reduce the computation in creating a new scale layer. Based on this principle, we implement 3D batch norm as shown in Fig. 4. For a 3D data cube, we flat in into a 2D plane image and apply the 2D batch norm in MKL2017 to get the result.

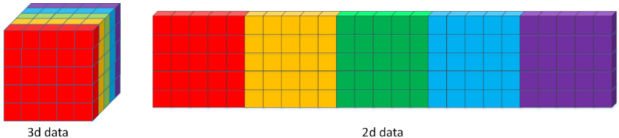

Figure 4: The principle of 3D batch norm with MKL2017.

## 3 RESULTS and DISCUSSION

### 3.1 Dataset

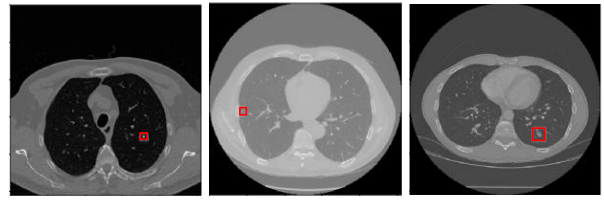

Figure 5: Examples of lung nodules in Luna'16 dataset.

We use Luna'16 dataset [14] for network training, which is based on the publicly available Lung Image Database Consortium (LIDC) [17] and the CT data form is DICOM. The dataset contain 888 CT scans and 1186 nodules in total. Some examples of lung nodules in Luna'16 dataset are shown in Fig. 5. Besides, the total Luna'16 dataset are divided into ten subsets, which are used for both training and testing CNN models. To allow easier reproducibility, the aforementioned lung nodule detection method is trained on the given subsets for 10-folds cross-validation.

### 3.2 Evaluation Measure

A candidate is considered to be a true positive if the distance between it and the nodule center is smaller than nodule radius. System performance is evaluated using Free Receiver Operating Characteristic (FROC) analysis, which considers both detection sensitivity and false positive rate per scan.

### 3.3 Lung Nodule Proposals Generation Results

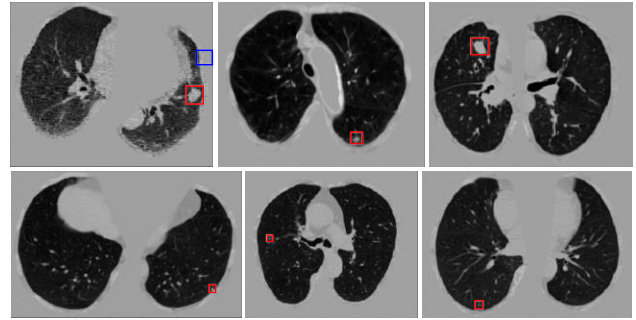

Figure 6: Some randomly chosen lung nodules generation results in Luna'16.

Some randomly chosen lung nodule proposals generation results by the aforementioned trained 3D CNN model are shown in Fig. 6, where regions in labeled bounding boxes are nodule proposals generation results, among which the red and blue bounding boxes are truly and falsely detection results, respectively. Results shown in the first and second rows of Fig. 6 illustrate that the constructed 3D CNN model can generate lung nodule proposals in large and small sizes, respectively, indicating the effectiveness of the adopted model.

We also quantitatively analyze the 3D CNN performance and obtain mean FROC value of 0.833 on Luna'16 dataset, ranking fourteen on Luna'16 competition (our method named as CCELargeCubeCnn) without any post-processing and false positive reduction procedures. Both the aforementioned experimental results indicate that the 3D CNN model achieves state-of-the-art lung nodule proposals generation performance.

### 3.4 Discussion of Data Resolution Impacts

As mentioned above, for raw CT scans, we firstly set sampling distance to convert them into 3D images with same interval and then crop a series of patches for network training. To analyze influences of data resolution on lung nodule proposals generation performances, we perform different sampling distances on the raw data and train the aforementioned 3D network with data in different resolutions. The lung nodule proposals generation results of different trained models are shown in Fig. 7, from which we can see that models trained by data in higher resolution achieves better performances, indicating that the CNN model can capture more distinguishing nodule features from high resolution data.

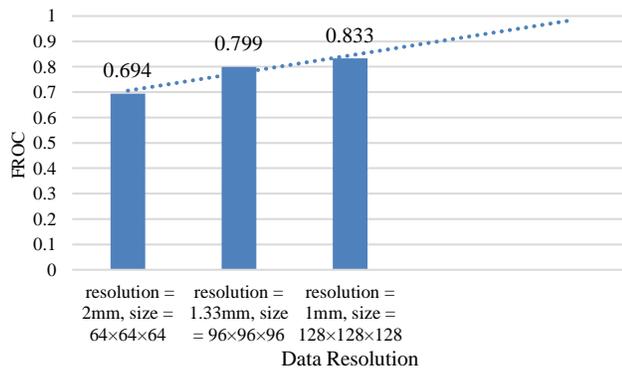

**Figure 7: Comparisons of nodule proposals generation performances with 3D models trained by data in different resolutions**

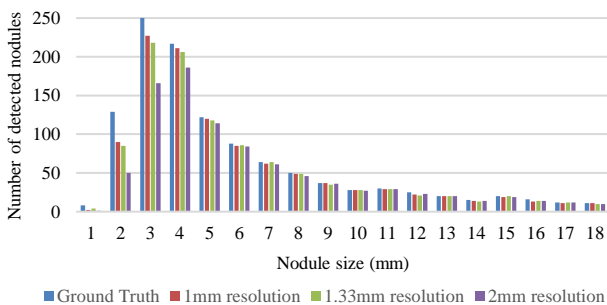

**Figure 8: Comparisons of small nodule proposal generation performances with 3D models trained by data in different resolutions**

To further find out the impact of data resolution, we analyze the proposal generation results of nodules in different sizes and show them in Fig. 8, from which we can see that, there are lots of small nodules in Luna'16 dataset and large nodules can be detected by all the models trained with data in different resolutions. However, models trained by higher resolution data can achieve better performance on generating small nodule proposals, because features of small nodules in higher resolution are more distinct and conspicuous.

### 3.5 Discussion of Memory Consumptions

In this paper, we apply 3D CNN model for lung nodule proposals generation and prefer high resolution data for network training, which consumes large memory. We implement the 3D CNN model on CPU and GPU, respectively, and analyze the memory consumption situations on these two platforms. The CPU we used is Intel Skylake with 384GB RAM, and the GPU we used is Maxwell with 12GB RAM.

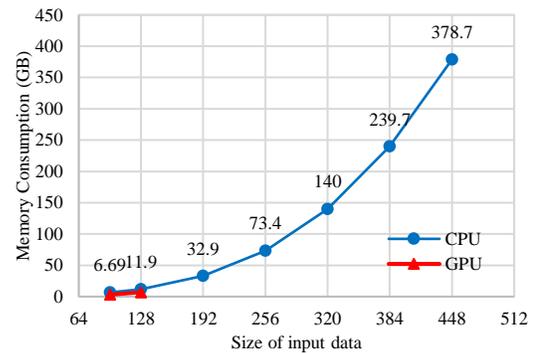

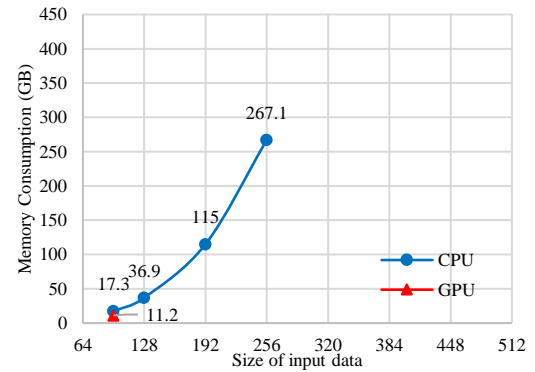

**Figure 9: Comparisons of memory consumptions with input data in different sizes on different platforms.**

Comparison of memory consumptions on different platforms with batch size set as 1 and 4 are shown in Fig. 9(a) and (b), respectively, from which we can see that the 3D model on GPU can receive the maximum input data of 128×128×128 pixels with batch size 1, which consumes 11.9GB RAM. However, the same model on CPU can receive the maximum input data of 448×448×448 pixels, which consumes 378.7GB RAM. When the batch size is set as 4, GPU can receive the maximum input data of 96×96×96 pixels, consuming memory of 11.2GB, while CPU receives the maximum input data of 256×256×256 pixels, consuming memory of 267.1GB. The aforementioned experimental results indicate that CPU with larger memory can receive higher resolution data and enables more possibilities of exploring larger memory required 3D applications.

### 3.6 Discussion of Performance Optimizations

As mentioned above, we implement the 3D computational functionalities and do some optimizations. In this section, we conduct some experiments to analyze the optimization performances.

**Table 2: Performances comparison results of different 3D convolutional units (ms)**

|  | Forward | Backward | Total |
| --- | --- | --- | --- |
| MKL-DNN | 1384.02 | 5742.72 | 7126.78 |
| SGEMM | 5280.38 | 8101.36 | 13381.73 |
| Performance ratio | 3.82 | 1.41 | 1.88 |

Time consumed by the total convolutional operations of the 3D CNN model using the aforementioned SGEMM and MKL-DNN based 3D convolutions respectively are illustrated in Table 2, from which we can see that the forward, backward and total consumption time is highly reduced after optimizations.

In addition, we compare performances of 3D batch norm with and without MKL2017, and show the results in Table 3, from which we can see that performance of batch norm optimized with MKL2017 is 3.42 times as higher as that of batch norm without optimization.

**Table 3: Performances comparison results of batch norm units with and without MKL-DNN engine (ms)**

|  | Forward | Backward | Total |
| --- | --- | --- | --- |
| With MKL2017 | 207.99 | 323.97 | 531.98 |
| Without MKL2017 | 1065.84 | 755.07 | 1820.91 |
| Performance ratio | 5.12 | 2.33 | 3.42 |

We also analyze the total time of training our model, the experimental results illustrate that it consumes about 61 hours to train the model for 100 epochs of one-fold dataset using our optimized Extended-Caffe, while time consumption without the aforementioned optimizations is about 293 hours.

## 4 CONCLUSION

Lung nodule proposals generation is the primary step of nodule detection which can help diagnose lung cancer early and promote the cure rate. In this paper, we construct a 3D CNN model for lung nodule proposals generation, which achieves state-of-the-art performance on Luna'16 competition with FROC equaling to 0.833. Based on this 3D CNN model, we discuss some key problems. Firstly, considering that lung nodules are usually in too small sizes and their features cannot be effectively extracted, we try to enlarge the nodule sizes relatively by enhancing training data resolutions and the experimental results indicating that high resolution input data can help improve the lung nodule proposals generation performance, while consuming large memory at the same time. Then we analyze memory consumption situations on different hardware platforms respectively and figure out that CPU architecture with large memory enables more possibilities to explore 3D applications. To meet the requirement of large memory, we implement the 3D CNN model on CPU architecture using our proposed Intel Extended-Caffe framework, which supports highly-efficient 3D computational operations and is released at https://github.com/extendedcaffe/extended-caffe.